\documentclass[conference]{IEEEtran}
\IEEEoverridecommandlockouts
\usepackage{amsmath,amssymb,amsfonts}
\usepackage{algorithmic}
\usepackage{graphicx}
\usepackage{textcomp}
\usepackage{xcolor}
\usepackage{booktabs, tabularx}
\usepackage{stfloats}
\usepackage{adjustbox}

\usepackage[colorlinks,pagebackref=false,citecolor=blue,bookmarks=false,hypertexnames=true]{hyperref}

\ifx00 
\newcommand{\sy}[1]{\textcolor{red}{$<$SY: #1$>$}}
\else
\newcommand{\sy}[1]{}
\fi

\usepackage[
maxnames=3,
maxbibnames=3,
maxcitenames=3,
minnames=3,
minbibnames=3,
mincitenames=3,
uniquelist=false
]{biblatex}
\begin{document}

\title{ViT-BEVSeg: A Hierarchical Transformer Network for Monocular Birds-Eye-View Segmentation}

\author{\IEEEauthorblockN{Pramit Dutta}
\IEEEauthorblockA{\textit{Dept of CS, Hamilton Institute} \\
\textit{Maynooth University}\\
Dublin, Ireland \\
pramit.dutta.2021@mumail.ie}
\and
\IEEEauthorblockN{Ganesh Sistu}
\IEEEauthorblockA{\textit{Computer Vision Department} \\
\textit{Valeo Vision Systems}\\
Galway, Ireland \\
ganesh.sistu@valeo.com}
\and
\IEEEauthorblockN{Senthil Yogamani}
\IEEEauthorblockA{\textit{Computer Vision Department} \\
\textit{Valeo Vision Systems}\\
Galway, Ireland \\
senthil.yogamani@valeo.com}
\and
\IEEEauthorblockN{\hspace{4.50cm}Edgar Galv\'an}
\IEEEauthorblockA{\hspace{4.50cm}\textit{Department of Computer Science} \\
\hspace{4.5cm}\textit{Maynooth University}\\
\hspace{4.5cm}Dublin, Ireland \\
\hspace{4.5cm}edgar.galvan@mu.ie}
\and
\IEEEauthorblockN{\hspace{-2cm}John McDonald}
\IEEEauthorblockA{\hspace{-2cm}\textit{Department of Computer Science} \\
\hspace{-2cm}\textit{Maynooth University}\\
\hspace{-2cm}Dublin, Ireland \\
\hspace{-2cm}john.mcdonald@mu.ie}
}

\maketitle

\begin{abstract}
Generating a detailed near-field perceptual model of the environment is an important and challenging problem in both self-driving vehicles and autonomous mobile robotics. A Bird’s Eye View (BEV) map, providing a panoptic representation, is a commonly used approach that provides a simplified 2D representation of the vehicle’s surroundings with accurate semantic level segmentation for many downstream tasks. Current state-of-the art approaches to generate BEV-maps employ a Convolutional Neural Network (CNN) backbone to create feature-maps which are passed through a spatial transformer to project the derived features onto the BEV coordinate frame.
In this paper, we evaluate the use of vision transformers (ViT) as a backbone architecture to generate BEV maps. Our network architecture, \textit{ViT-BEVSeg}, employs standard vision transformers to generate a multi-scale representation of the input image. The resulting representation is then provided as an input to a spatial transformer decoder module which outputs segmentation maps in the BEV grid. We evaluate our approach on the nuScenes dataset demonstrating a considerable improvement in the performance relative to state-of-the-art approaches. Code is available at ~\url{https://github.com/robotvisionmu/ViT-BEVSeg}.
\end{abstract}


\section{Introduction}\label{introduction}
\begin{figure}[tb]
\centerline{\includegraphics[width=0.5\textwidth]{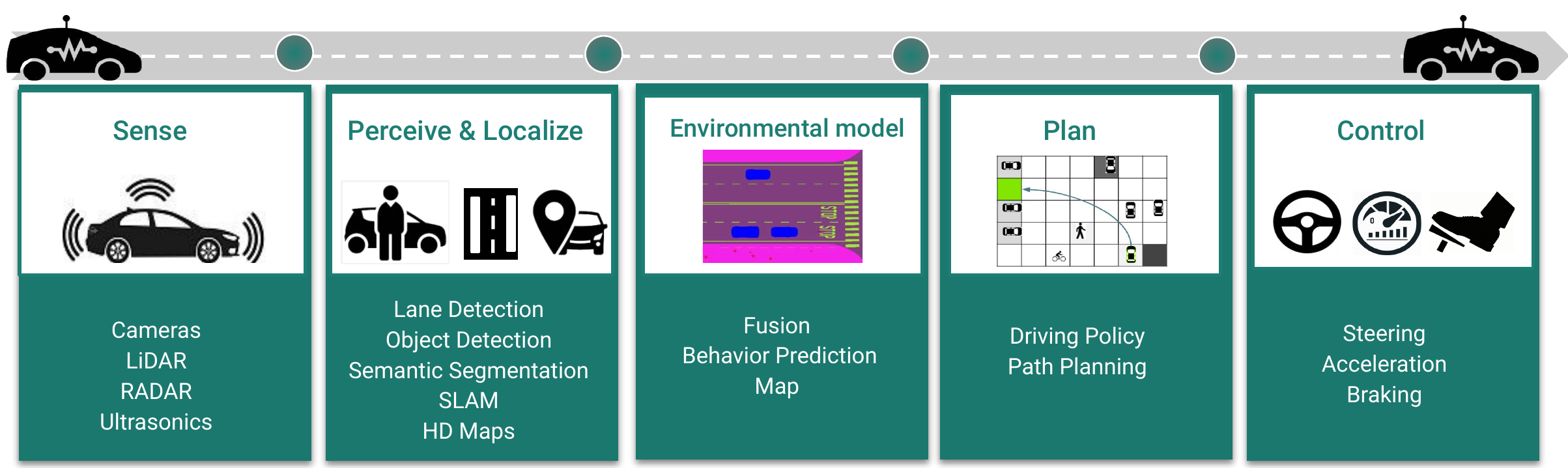}}
\caption{Standard components in a modern autonomous driving systems pipeline listing the various tasks. Environmental model is represented using a simplified Bird's eye view (BEV) grid.}
\label{fig1}
\end{figure}

Autonomous mobility systems such as self-driving vehicles, mobile robotics and advanced driver assistance systems have seen significant advances over recent years. This has been facilitated by progress in machine learning coupled with sophisticated neural-network models trained on a variety of complex large scale data-sets. As described in~\cite{shadrin2019analytical}, mobility systems can be classified based on the amount of automation provided by the sensors and data-processing stack embedded within the vehicle's control architecture, see Fig.~\ref{fig1}. The central part of the autonomous driving pipeline is the environmental model in which the output of the perception are represented. Subsequent downstream tasks such as trajectory prediction and path planning use this representation as their input.

A common approach for research in mobile robotic platforms and autonomous vehicles is to employ active LIDAR systems to capture three-dimensional (3D) point clouds of their surroundings in real-time~\cite{capellier2018evidential,lu2019l3}. However, such sensors suffer from various limitations including sparse measurements, low scanning rates and high-cost, each of which has impeded their adoption in commercial vehicles~\cite{dwivedi2021bird}. A more generalized solution is to use multiple passive monocular cameras to capture the environment whereby the resulting data is interpreted using computer vision and machine learning techniques for tasks such as object detection~\cite{rashed2021generalized}, distance estimation~\cite{kumar2020unrectdepthnet}, soiling detection~\cite{uricar2021let, uricar2019desoiling}, and multi-task learning~\cite{chennupati2019auxnet, sistu2019real}. These perception tasks are often solved directly in image-space which is defined relative to the associated camera's frame of reference. An intuitive and aggregated description of the vehicle's surroundings can be generated by mapping the resulting image features to the vehicle's ego frame of reference.
Bird's Eye View (BEV) maps provide one such simplified representation which are commonly used as an environmental model for automated driving systems. BEV maps capture the spatial configuration of the vehicle's surroundings and provide a top-down grid with semantic information.

Recently, a number of deep learning-based approaches have been proposed to directly predict BEV maps from input images, including  VED~\cite{lu2019monocular}, VPN~\cite{pan2020cross}, PON~\cite{roddick2020predicting}, Lift-Splat-shoot~\cite{philion2020lift}, and Lifted-2D semantic segmentation~\cite{dwivedi2021bird}. In each of these approaches, the pipeline used to generate the BEV map utilises a Convolutional Neural Networks (CNN) as the backbone to extract one or two-dimensional features from the input images and then project these features to the BEV plane using a spatial transformation layer~\cite{jaderberg2015spatial}. The semantic features are then aggregated using various network models to generate an occupancy grid map~\cite{elfes1990stochastic}. 

In recent years, self-attention based architectures through vision transformers~\cite{dosovitskiy2020image,liu2021swin} have been shown to achieve considerable improvement in performance over CNN based architectures~\cite{he2016deep,tan2019efficientnet} in computer vision tasks such as classification, semantic segmentation, instance segmentation, and more. A comprehensive review on vision transformers can be found in~\cite{han2020survey}. 

In this work, we make use of a vision transformer as an alternative backbone to develop a multi-resolution encoder-decoder architecture to estimate BEV occupancy grid maps from monocular front view images. We reconfigure the vision transformer~\cite{dosovitskiy2020image} in a similar manner to~\cite{ranftl2021vision} to extract a hierarchical embedded representation of the input image and generate assembled feature maps at different resolutions (as described in Section~\ref{networkarchitecture}). The feature-maps are then projected to the BEV plane using a dense transformer layer~\cite{roddick2020predicting}.  We perform end-to-end model training on the publicly available large scale nuScenes dataset~\cite{caesar2020nuscenes}. 

The main contributions of the work are:
\begin{enumerate}
    \item We present a BEV map estimation architecture that employs a vision transformer network as the feature encoder backbone.
    \item We combine a series of vision transformer blocks in a hierarchical configuration to compute and fuse representations across multiple spatial scales.
    \item Our architecture can be trained end-to-end to generate semantic occupancy maps in birds-eye-view.
    \item We train and evaluate the network architecture using the nuScenes  dataset demonstrating a significant improvement over similar models with CNN backbones. 
\end{enumerate}
 
The remaining sections of this paper are organized as follows.
Section~\ref{related} details the related work pertaining to BEV maps generation and vision transformers. Section~\ref{networkarchitecture} outlines the proposed network architecture. Section ~\ref{experiment} presents the experimental setup and discusses the results obtained by our proposed approach. Section~\ref{sec:conclusions} provides concluding comments and discusses future directions for the research.

\section{Related Work}\label{related}
 
\textbf{BEV Map Generation from Monocular Images:} Generating BEV maps with semantic segmentation using monocular images is a complex task that requires multi-level processing of the input images. In general, the pipeline can be divided into three stages: (1) Extracting image features to generate feature maps, (2) Projecting feature maps from front-view to top-down (BEV) view, and (3) Aggregating various semantic representations onto a map with respect to the vehicle's egocentric reference frame. Various approaches have been reported in the literature for each of these stages.

As in typical computer vision problems of semantic and/or instance segmentation, an encoder-decoder network is employed for BEV map generation from monocular images. The encoder reduces the size of activations (features), using convolution and pooling layers, to generate a latent state representation. The decoder network uses de-convolution and un-pooling layers to enlarge the activations to map feature pixels to output categories. This architecture allows end-to-end training of the resulting network~\cite{hao2020brief,lateef2019survey}. The encoder in most models for BEV map generation uses a CNN based architecture such as ResNet~\cite{roddick2020predicting,dwivedi2021bird} or EfficientNet~\cite{gosala2022bird} as the backbone. The features are extracted at different image resolutions and hierarchically merged by up-sampling. Such a pyramid representation helps to augment fine-grained spatial configurations from lower layers of the pyramid to high resolution features \cite{lin2017feature}. The second stage of the BEV map generation pipeline involves projecting image features from the input front-view to the BEV top-view. In~\cite{lin2012vision} the authors employed an image to ground-plane homography to apply an inverse-perspective mapping to achieve this spatial transformation. In~\cite{philion2020lift} the authors un-project the features-maps onto a volumetric lattice to represent each image pixel in a pseudo point cloud matrix and use a pillar-pooling method, inspired from point-cloud processing to estimate a categorical depth distribution for a given camera rig. The work in \cite{roddick2020predicting} is most similar to our work where the authors use a pyramid dense transformer decoder to project each output feature-map from the encoder to top-view for a subset of depth values. More recently, \cite{gosala2022bird} train two distinct spatial transformers in parallel to independently map features from vertical and flat regions from front view to top view.

\textbf{Vision Transformers as Encoders:} Inspired from the success in the Natural Language Processing (NLP) domain~\cite{vaswani2017attention}, transformer networks for vision have shown considerable improvement over CNN architectures in many computer vision tasks. The vision transformers use a multi-head self attention mechanism~\cite{vaswani2017attention}. The input image is split into multiple patches and flattened to an embedding dimension. The patches are provided with positional encoding~\cite{dosovitskiy2020image} and processed through the transformer blocks, discussed further in section~\ref{VTE}. Vision transformers are pre-trained over large datasets with a supervised classification task and can be fine-tuned to perform downstream tasks. Variations of the original vision transformer have been proposed to tackle specific problems of data efficient training~\cite{touvron2021training}, image size scalability~\cite{liu2021swin} as well as task agnostic models for dense prediction~\cite{wang2021pyramid,ranftl2021vision}, semantic segmentation~\cite{xie2021segformer}, and object detection~\cite{zhu2020deformable}. Exhaustive surveys on the current transformer models, their variations and applications to the domain of computer vision can be found in~\cite{han2020survey,khan2021transformers}. 

In our work, we present a BEV map estimation architecture using vision transformer network encoder backbone. We primarily focus on the original vision transformer architecture~\cite{dosovitskiy2020image} and its two variations, ViT-Base-16 and ViT-Large-16, as a backbone. By applying suitable modifications, to generate hierarchical feature maps, we achieve state of the art performance on multiple object and background classes when evaluated on the nuScenes dataset.

\begin{figure*}[t]
\centering
\includegraphics[width=0.90\textwidth]{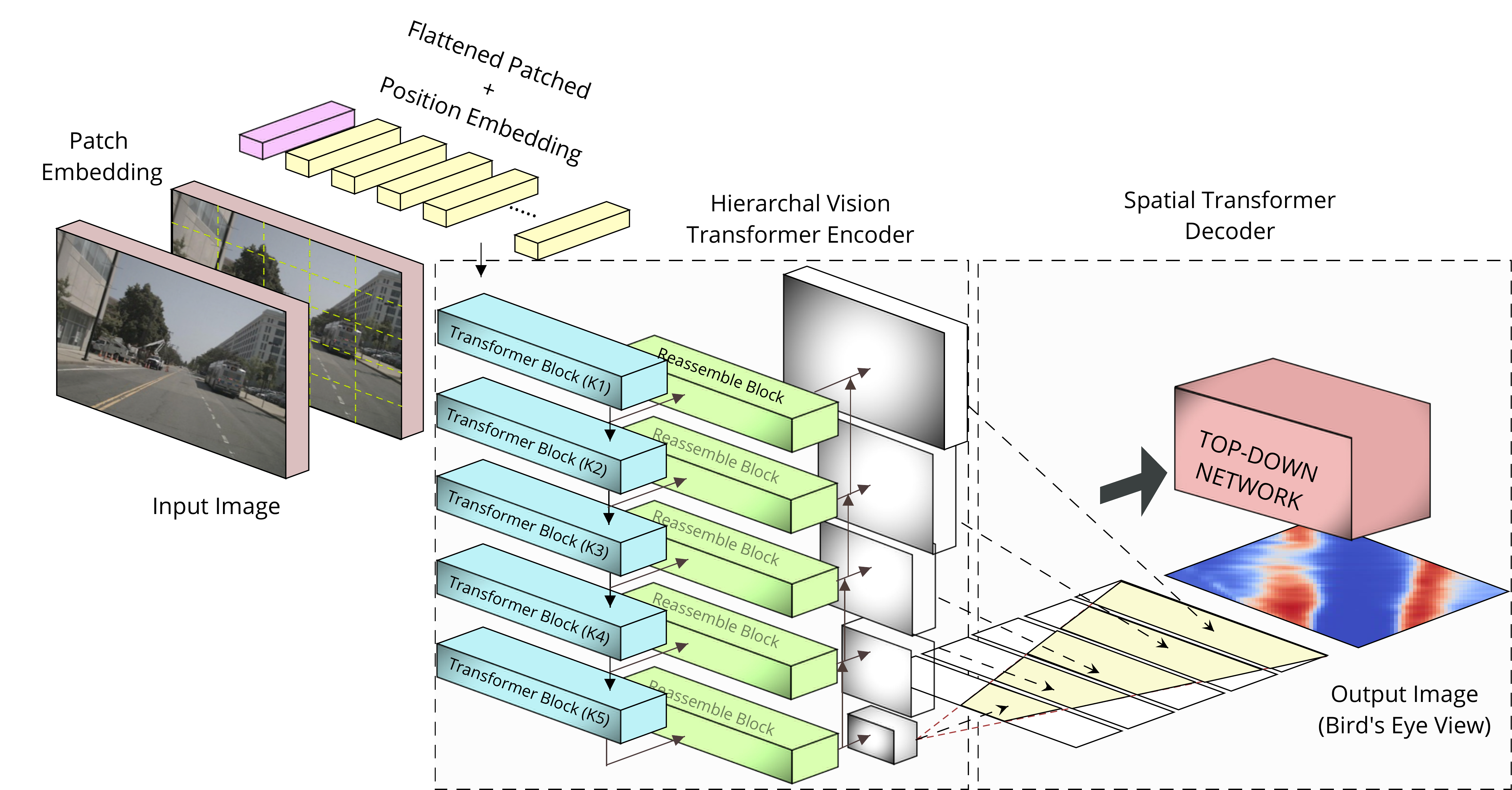}
\caption{An overview of the proposed model ViT-BEVSeg architecture: (a) The input image is divided into patches, flattened and projected to embeddings. (b) The embeddings are processed using a standard vision transformer blocks and features are extracted from various blocks (c) The features are reassembled into image like configuration at different resolutions using a pyramid architecture (d) A spatial transformer layer projects the image features to BEV to generate semantic occupancy grid maps.}
\label{architecture}
\end{figure*}

\section{Proposed Method}\label{networkarchitecture}

In this section, we provide an overview of the proposed network architecture to generate BEV maps using vision transformers as a backbone. As shown in Fig.~\ref{architecture}, the network retains the encoder-decoder configuration as used in previous works allowing end-to-end training of the model.

\subsection{Hierarchical Vision Transformer Encoder}\label{VTE} 

The encoder network comprises of a vision transformer backbone that takes as input a front-view image $x\in\mathbb{R}^{C \times H \times W}$. The image is split into $N_p$ image patches with a patch resolution of $p \times p$ pixels. Each patch is flattened and mapped to an embedded or latent feature space with dimension $D$. The patch embedding is a trainable linear projection, the dimensions of which remain constant throughout the transformer blocks. Each of these embedded feature patches is referred to as a \emph{token}. A non-grounded additional token $t_0$ is appended that provides the global image representation for classification. This approach is similar to using the \emph{cls\textunderscore token} in transformers for NLP. The tokens have a linear correspondence with the embedded patches thus the spatial resolution is maintained through the transformer stages. The final input for the transformer can thus be represented as \(z^0 = [t_0^0;t_1^0,..,t_{N_p}^0]\), and the output from transformer blocks can be given as $t_n^\kappa \in \mathbb{R}^{(N_p + 1) \times D}$ and $\kappa\in K$. $K$ represents the maximum number of transformer blocks in the encoder. Each transformer block is comprised of alternating layers of multi-headed self-attention (MHSA), MLP blocks and layer normalisation layers (LN) \cite{dosovitskiy2020image}. Attention is achieved using Eq.~\ref{eq:attention} by creating `query, key and value' learnable parameters over the tokens. 

\begin{equation}
    Attention(Q,K,V) = softmax \left(\frac{QK^T}{\sqrt{d_k}}\right)V
    \label{eq:attention}
\end{equation} As all embedded patches attend over each other, this inherently provides a global receptive field, i.e. over the entire image, thus significantly improving the performance over CNN networks when trained on large datasets~\cite{dosovitskiy2020image}.

 In this work, we use two transformer models, ViT-Base-16 and ViT-Large-16 \cite{dosovitskiy2020image} as the model backbone. Feature embeddings are extracted from five different blocks within the transformer encoder. These linear embeddings are reassembled into image shaped features at different resolutions. As in a typical pyramid network \cite{lin2017feature}, feature maps from the initial layers are assembled at higher resolutions than those derived from the deeper layers. The maps are progressively up-sampled and added to create the final set of feature maps to be passed onto the decoder network. The additional classification token $t_0$ is not grounded to the input image and thus can be ignored during the reassembly procedure. However, based on the ablation study provided in \cite{ranftl2021vision}, to handle the $t_0$ token, we concatenate it to each of the patch embeddings and re-project it to $D$ to retain the information learned, prior to performing the reassembly operation. The output dimension of each of the feature maps from the encoder is $I\in\mathbb{R}^{\hat{D} \times \frac{H}{l_i} \times \frac{W}{l_i}}$, where $\hat{D}$ is the final projected feature dimensions which is constant across all resolutions. Tables~\ref{table:1} and~\ref{table:2} give details of the parameters used in our model and described above. 
\begin{table}[htb]
\centering
\caption{Encoder Transformer Parameters}
\label{table:1}
\begin{tabular}{p{0.35\linewidth}|p{0.2\linewidth}|p{0.2\linewidth}}
\hline
\textbf{Model} & \textbf{ViT(B)-BEVSeg} &\textbf{ViT(L)-BEVSeg}  \\
\hline
\textbf{Img Resolution $(CHW)$} & $3 \times 384 \times 384$ & $3 \times 384 \times 384$ \\
\textbf{Patch Res. $(P)$} & $16$ & $16$ \\
\textbf{Embed. Di. $(D)$} & $768$ & $1024$ \\
\textbf{Num. of Blocks $(K)$} & $12$ & $24$ \\
\hline
\end{tabular}
\end{table}

\begin{table}[htb]
\centering
\caption{Feature Reassembly Parameters}
\label{table:2}
\begin{tabular}{p{0.3\linewidth}|p{0.25\linewidth}|p{0.25\linewidth}}
\hline
\textbf{Model} & \textbf{ViT(B)-BEVSeg} &\textbf{ViT(L)-BEVSeg}  \\
\hline
\textbf{Output Blocks $(\kappa_i)$} & ${3,5,7,9,12}$ & ${5,10,15,20,24}$ \\
\textbf{Reassembly Res. $(l_i)$} & ${\frac{1}{4},\frac{1}{8},\frac{1}{16},\frac{1}{32},\frac{1}{64}}$ & ${\frac{1}{4},\frac{1}{8},\frac{1}{16},\frac{1}{32},\frac{1}{64}}$ \\
\textbf{Projected Dim. $(\hat{D})$} & $256$ & $256$ \\
\hline
\end{tabular}
\end{table}

\subsection{Spatial Transformer Decoder}
The feature maps are transformed from front view image space to orthographic BEV space, using a pyramid of basic spatial transformers inspired from \cite{roddick2020predicting}. The feature map with dimensions $I\in\mathbb{R}^{\hat{D} \times \frac{H}{l_i} \times \frac{W}{l_i}}$ is transformed to a BEV image with dimensions $I_{BEV} \in \mathbb{R}^{\hat{D} \times \frac{Z}{j} \times \frac{W}{l_i}}$ using three operations. In the first operation, the height of the feature map is collapsed to a bottleneck of size $B$. The convolution network projects the bottleneck features into polar feature of dimensions  $\hat{D} \times Z \times W$. Lastly, the polar features are projected to a Cartesian plane using the camera intrinsic parameters and horizontal offset. A five level pyramid of such transformers acts on each of the outputs from the encoder to map the features for a subset of depth values. The output semantic grid map is thus constructed by concatenating the output from each of the transformers in the pyramid. 

\subsection{Loss Function}
Generally for semantic segmentation problems a Binary Cross Entropy (BCE) loss is used to train the network model and create semantic occupancy probabilities to match the ground-truth. When the data-set have a definitive class imbalance both in the number of semantic instances and pixel size variation, we can introduce prior bias to weigh the occupancy probability of classes. Furthermore, to include instances with higher uncertainty (regions with occlusions or uncertain terrain), an additional parameter is introduced into the loss function that deliberately increases the occupancy predictions for ambiguous regions of the feature map. Thus for $c$ classes, with class probabilities $p = [p^1,p^2,...,p^c]$ with $p^i \in [0,1)$, ground truth $\hat{p}^i$,  class weight $w^c$, and uncertainty factor of $\lambda$. The loss function is given as,
\begin{equation}
\begin{split}
    Loss = \sum\limits_{i=1}^c w^i \hat{p}^i\log(p^i) + (1-w^i)(1-\hat{p}^i)\log(1-p^i) \\+ \lambda (1 - p^i \log_{2}p^i)
    \label{eq:loss}
\end{split}
\end{equation}

\begin{table*}[ht]
\caption{IoU Scores (\%) on NuScenes Dataset. The reported numbers are from the corresponding original papers. CS Mean refers to Average of Classes present in Cityscapes Dataset shown by*. \\ViT(B)-BEVSEG and ViT(L)-BEVSEG are our proposed methods using the ViT-Base-16 and ViT-Large-16 as the Backbone. Abbreviations: PC-Pedestrian Crossing; WW-Walkway; CP-Car park; CV-Construction Vehicle; Ped-Pedestrians; 2W-Two wheeler; TC-Traffic Cone; TB-Traffic Barriers.}
\label{table:results}
\begin{adjustbox}{width=\textwidth}
\begin{tabular}{@{}ccccccccccccccc|cc@{}}
\toprule
\textbf{Model} & \textbf{Drivable*} & \textbf{PC} & \textbf{WW*} & \textbf{CP} & \textbf{Car*} & \textbf{Truck} & \textbf{Bus*} & \textbf{Trailer} & \textbf{CV} & \textbf{Ped*} & \textbf{2W*} & \textbf{Cycle*} & \textbf{TC} & \textbf{TB} & \textbf{Mean} & \textbf{CS Mean} \\
\toprule
\textbf{IPM} & 40.1 & - & 14.0 & - & 4.9 & - & 3.0 & - & - & 0.6 & 0.8 & 0.2 & - & - & - & 9.1 \\
\textbf{VED}~\cite{lu2019monocular} & 54.7 & 12.0 & 20.7 & 13.5 & 8.8 & 0.2 & 0.0 & 7.4 & 0.0 & 0.0 & 0.0 & 0.0 & 0.0 & 4.0 & 8.7 & 12.0 \\
\textbf{VPN}~\cite{pan2020cross} & 58.0 & 27.3 & 29.4 & 12.9 & 25.5 & 17.3 & 20.0 & 16.6 & 4.9 & 7.1 & 5.6 & 4.4 & 4.6 & 10.8 & 17.5 & 21.4 \\
\textbf{PON}~\cite{roddick2020predicting} & 60.4 & 28.0 & 31.0 & 18.4 & 24.7 & 16.8 & 20.8 & 16.6 & 12.3 & \textbf{8.2} & 7.0 & \textbf{9.4} & 5.7 & 8.1 & 19.1 & 23.1 \\
\textbf{OFT}~\cite{roddick2018orthographic} & 62.4 & 30.9 & 34.5 & 23.5 & 34.7 & 17.4 & 23.2 & 18.2 & 3.7 & 1.2 & 6.6 & 4.6 & 1.1 & 12.9 & 19.6 & 23.9 \\
\textbf{2D-Lift}~\cite{dwivedi2021bird} & 62.3 & 31.8 & 37.3 & 25.2 & 37.4 & 18.7 & 24.8 & 16.4 & 4.7 & 3.4 & 7.9 & 7.2 & 3.9 & \textbf{13.6} & 21.0 & 25.8 \\
\textbf{2D-Lift + Aux Task}~\cite{dwivedi2021bird} & 61.1 & 33.5 & 37.8 & 25.4 & \textbf{37.8} & \textbf{20.4} & \textbf{31.8} & 14.2 & 2.7 & 5.9 & \textbf{10.5} & 6.69 & \textbf{7.57} & 13.4 & 22.1 & 27.4 \\
\toprule
    \textbf{ViT(B)-BEVSeg} & \textbf{80.9} & \textbf{42.7} & \textbf{57.3} & \textbf{57.2} & 25.6 & 18.7 & 27.1 & \textbf{25.2} & \textbf{17.4} & 5.7 & 8.0 & 4.3 & 4.2 & 12.9 & \textbf{27.6} & \textbf{29.8} \\
\textbf{ViT(L)-BEVSeg} & 74.8 & 33.1 & 48.2 & 48.5 & 22.5 & 13.6 & 17.6 & 20.5 & 13.1 & 4.8 & 6.2 & 3.7 & 3.5 & 11.0 & 22.95 & 25.4\\
\toprule
\end{tabular}
\end{adjustbox}
\end{table*}

\section{Experimental Evaluation}\label{experiment}

\subsection{Model Setup}
To evaluate our proposed model architecture we use two baseline vision transformer models namely, ViT-Base-16 and ViT-Large-16 as backbones. Table~\ref{table:2} provide details of the blocks where the features are extracted and corresponding resolutions at which the features are assembled. All our models are implemented in PyTorch. We use the ViT-(base/large) pre-trained weights from the PyTorch Timm~\cite{rw2019timm} library to initialize the training. The input images are scaled to match the transformer input resolution of $384\times384$ pixels. The training uses a SGD optimizer with initial learning rate of $0.05$ and weight decay of $0.001$. The models are trained using a single NVidia GTX1080Ti GPU with a batch size of $4$ for $50$ epochs in both cases. As in most semantic segmentation problems, we use the class-wise intersection over union (IoU $\%$) score to evaluate the performance of the model during training and validation.

\subsection{Dataset Setup}
The models are trained and tested on the publicly available nuScenes dataset \cite{caesar2020nuscenes}. The dataset, in its raw form, consists of 1000 driving scenes each of which is 20 seconds in duration. The complete dataset comprises of approximately 1.4 million images with calibrated camera parameters and associated LIDAR data with 3D annotations of bounding boxes. In total, there are 23 object class annotations (e.g. cars, trucks, motorcycles, etc.) and 9 background class annotations (driveable surface, sidewalk, pedestrian crossing, etc.). We use 14 classes to train the model: 10 object classes and 4 background classes. While creating the ground truth BEV semantic maps we restrict the depth extent for each camera at 40m with a horizontal field at 20m on either side of the camera. We use the default train-validation split as defined by the nuScenes dataset which is consistent with the previous reported works. The resultant ground truth image renders binary images with 2D boxes for each class mapped onto an occupancy grid map in the BEV plane. 

\begin{figure*}[]
\centering
\includegraphics[width=\textwidth]{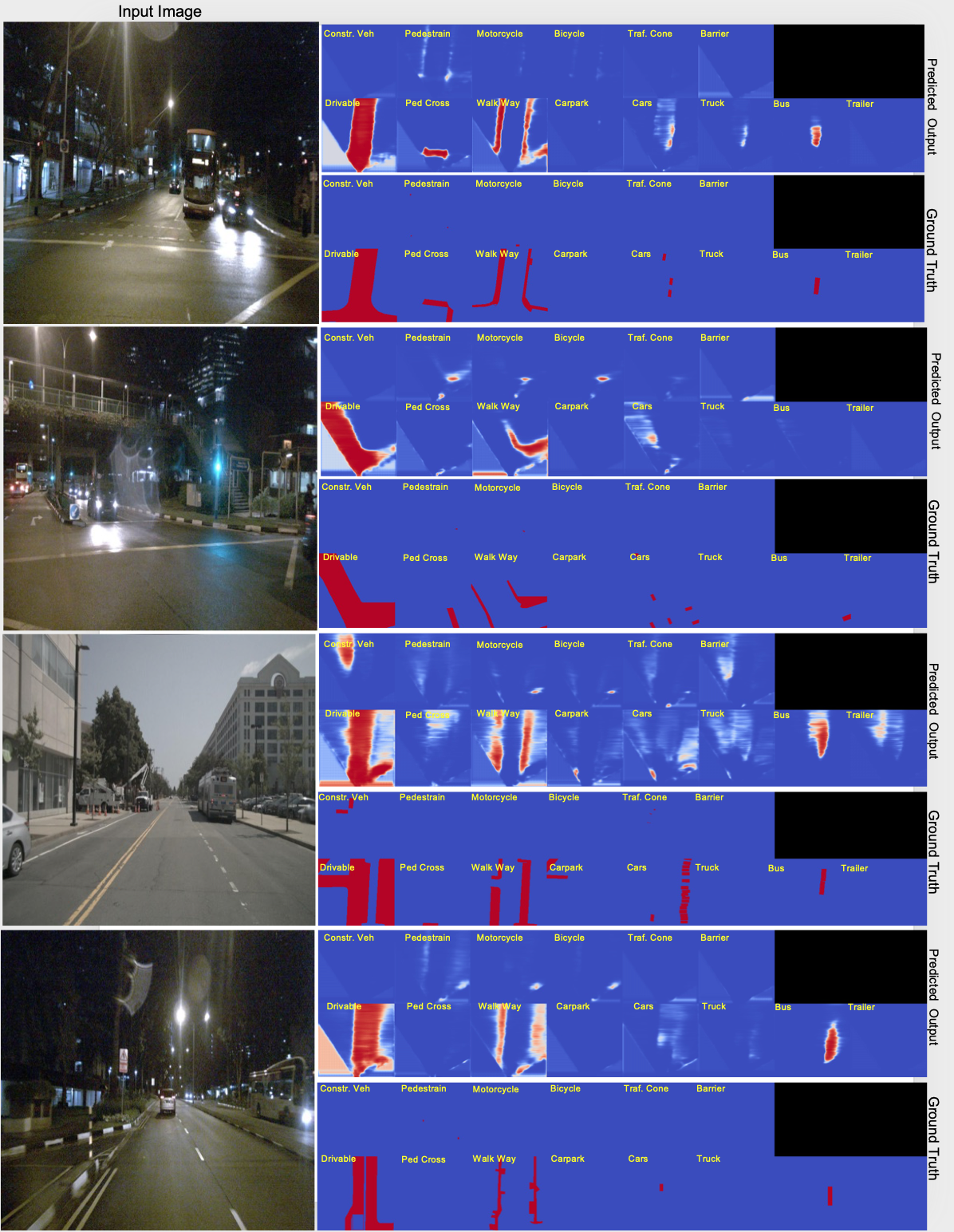}
\caption{Qualitative results showing class-wise predictions of our model against the ground truth for each input.}
\label{fig:results}
\end{figure*}

\subsection{Results}
We compare the performance of our models using the IoU scores on classes and mean with the results presented for the current baseline and competing works. A sample of qualitative results across both day and night time scenarios are shown in Fig.~\ref{fig:results}. Table~\ref{table:results} shows the comparative results on all the $14$ classes. Our model with a backbone of hierarchical vision transformer encoders achieves state-of-the-art performance in six of the fourteen classes. Our model improves upon the mean IoU score and the Cityscapes \cite{Cordts2016Cityscapes} mean IoU score by $24.9\%$ and $8.7\%$, respectively. We also note that our model does better in classes: truck, bus and two-wheeler, when compared to the baseline model as reported in~\cite{dwivedi2021bird}. However,~\cite{dwivedi2021bird} also demonstrate that when training with perspective semantic segmentation as an auxiliary task, the model performance improves in certain classes. We identify to explore this training strategy and its impact in our future work. The proposed model, however, under performs in certain categories: car, pedestrians, bicycle, barrier and traffic-cone. On further evaluation we notice that these categories can be attributed as small sized objects as compared the other classes. As we do not introduce random crop augmentation in the training data-pipeline and scale the input image size to fit the vision transformer model, our training images are significantly down sampled from approximately $800\times600$ pixels as used in some of the previous works to $384\times384$. This could be attributed as a possible explanation for a low performance of the model for smaller sized object categories.

\section{Conclusions}\label{sec:conclusions}
In this paper, we address the problem of BEV map generation with semantic segmentation using a vision transformer as a backbone for the encoder. This is in contrast to prior work that only used CNN based backbones. The vision transformer is reconfigured with additional modules to extract feature maps from different levels of the transformer blocks and create a hierarchical assembly network to produce a pyramid like representations at different image resolutions. A decoder consisting of a spatial transformation networks converts the front view features to top view to generate a semantic occupancy grid map. After the training the model on nuScences dataset, our model architecture considerably improves on the IoU metric over the current models in most class categories. We identify that the probable cause for under performance in certain classes, that may be considered as small sized objects as compared to other classes, can be attributed to the scaled input image size used for the vision transformer. As vision transformers are becoming more popular, newer models have been proposed to tackle the problem of image scaling and semantic segmentation. We envisage to extend our current work in three specific aspects. Firstly, we will extend our study for BEV map generation using different vision transformers such as SWIN transformer~\cite{liu2021swin}, Segformer~\cite{xie2021segformer} etc. Secondly, we would like to extend our model to perform BEV map generation on input images from fish-eye cameras, given their prevalence in autonomous driving platforms due to their larger field of view. Lastly, we intend to investigate the performance improvement by addition of an auxiliary tasks such as perspective semantic segmentation within the current network.

\section*{Acknowledgments}

        This work has emanated from research conducted with the financial support of Science Foundation Ireland (SFI) under Grant Number SFI 18/CRT/6049 and in part by a research grant from Science Foundation Ireland (SFI) under Grant Number 16/RI/3399. 

{
\renewcommand*{\bibfont}{\small}
\printbibliography

@inproceedings{roddick2020predicting,
  title={Predicting semantic map representations from images using pyramid occupancy networks},
  author={Roddick, Thomas and Cipolla, Roberto},
  booktitle={Proceedings of the IEEE/CVF Conference on Computer Vision and Pattern Recognition},
  pages={11138--11147},
  year={2020}
}

@inproceedings{philion2020lift,
  title={Lift, splat, shoot: Encoding images from arbitrary camera rigs by implicitly unprojecting to 3d},
  author={Philion, Jonah and Fidler, Sanja},
  booktitle={European Conference on Computer Vision},
  pages={194--210},
  year={2020},
  organization={Springer}
}

@article{dosovitskiy2020image,
  title={An image is worth 16x16 words: Transformers for image recognition at scale},
  author={Dosovitskiy, Alexey and Beyer, Lucas and Kolesnikov, Alexander and Weissenborn, Dirk and Zhai, Xiaohua and Unterthiner, Thomas and Dehghani, Mostafa and Minderer, Matthias and Heigold, Georg and Gelly, Sylvain and others},
  journal={arXiv preprint arXiv:2010.11929},
  year={2020}
}

@inproceedings{liu2021swin,
  title={Swin transformer: Hierarchical vision transformer using shifted windows},
  author={Liu, Ze and Lin, Yutong and Cao, Yue and Hu, Han and Wei, Yixuan and Zhang, Zheng and Lin, Stephen and Guo, Baining},
  booktitle={Proceedings of the IEEE/CVF International Conference on Computer Vision},
  pages={10012--10022},
  year={2021}
}

@article{han2020survey,
  title={A survey on visual transformer},
  author={Han, Kai and Wang, Yunhe and Chen, Hanting and Chen, Xinghao and Guo, Jianyuan and Liu, Zhenhua and Tang, Yehui and Xiao, An and Xu, Chunjing and Xu, Yixing and others},
  journal={arXiv e-prints},
  pages={arXiv--2012},
  year={2020}
}

@article{khan2021transformers,
  title={Transformers in vision: A survey},
  author={Khan, Salman and Naseer, Muzammal and Hayat, Munawar and Zamir, Syed Waqas and Khan, Fahad Shahbaz and Shah, Mubarak},
  journal={ACM Computing Surveys (CSUR)},
  year={2021},
  publisher={ACM New York, NY}
}

@inproceedings{wang2021pyramid,
  title={Pyramid vision transformer: A versatile backbone for dense prediction without convolutions},
  author={Wang, Wenhai and Xie, Enze and Li, Xiang and Fan, Deng-Ping and Song, Kaitao and Liang, Ding and Lu, Tong and Luo, Ping and Shao, Ling},
  booktitle={Proceedings of the IEEE/CVF International Conference on Computer Vision},
  pages={568--578},
  year={2021}
}

@article{lu2019monocular,
  title={Monocular semantic occupancy grid mapping with convolutional variational encoder--decoder networks},
  author={Lu, Chenyang and van de Molengraft, Marinus Jacobus Gerardus and Dubbelman, Gijs},
  journal={IEEE Robotics and Automation Letters},
  volume={4},
  number={2},
  pages={445--452},
  year={2019},
  publisher={IEEE}
}

@inproceedings{elfes1990stochastic,
  title={A stochastic spatial representation for Active Robot Perception},
  author={Elfes, A},
  booktitle={Proc. on the sixth Conference on Uncertainty and AI, AAAI, Cambridge, MA},
  year={1990}
}

@article{dwivedi2021bird,
  title={Bird’s Eye View Segmentation Using Lifted 2D Semantic Features},
  author={Dwivedi, Isht and Malla, Srikanth and Chen, Yi-Ting and Dariush, Behzad},
  year={2021}
}

@inproceedings{ranftl2021vision,
  title={Vision transformers for dense prediction},
  author={Ranftl, Ren{\'e} and Bochkovskiy, Alexey and Koltun, Vladlen},
  booktitle={Proceedings of the IEEE/CVF International Conference on Computer Vision},
  pages={12179--12188},
  year={2021}
}

@article{vaswani2017attention,
  title={Attention is all you need},
  author={Vaswani, Ashish and Shazeer, Noam and Parmar, Niki and Uszkoreit, Jakob and Jones, Llion and Gomez, Aidan N and Kaiser, {\L}ukasz and Polosukhin, Illia},
  journal={Advances in neural information processing systems},
  volume={30},
  year={2017}
}

@article{gosala2022bird,
  title={Bird's-Eye-View Panoptic Segmentation Using Monocular Frontal View Images},
  author={Gosala, Nikhil and Valada, Abhinav},
  journal={IEEE Robotics and Automation Letters},
  year={2022},
  publisher={IEEE}
}

@article{lin2012vision,
  title={A vision based top-view transformation model for a vehicle parking assistant},
  author={Lin, Chien-Chuan and Wang, Ming-Shi},
  journal={Sensors},
  volume={12},
  number={4},
  pages={4431--4446},
  year={2012},
  publisher={Molecular Diversity Preservation International}
}

@inproceedings{lin2017feature,
  title={Feature pyramid networks for object detection},
  author={Lin, Tsung-Yi and Doll{\'a}r, Piotr and Girshick, Ross and He, Kaiming and Hariharan, Bharath and Belongie, Serge},
  booktitle={Proceedings of the IEEE conference on computer vision and pattern recognition},
  pages={2117--2125},
  year={2017}
}

@article{pan2020cross,
  title={Cross-view semantic segmentation for sensing surroundings},
  author={Pan, Bowen and Sun, Jiankai and Leung, Ho Yin Tiga and Andonian, Alex and Zhou, Bolei},
  journal={IEEE Robotics and Automation Letters},
  volume={5},
  number={3},
  pages={4867--4873},
  year={2020},
  publisher={IEEE}
}

@inproceedings{caesar2020nuscenes,
  title={nuscenes: A multimodal dataset for autonomous driving},
  author={Caesar, Holger and Bankiti, Varun and Lang, Alex H and Vora, Sourabh and Liong, Venice Erin and Xu, Qiang and Krishnan, Anush and Pan, Yu and Baldan, Giancarlo and Beijbom, Oscar},
  booktitle={Proceedings of the IEEE/CVF conference on computer vision and pattern recognition},
  pages={11621--11631},
  year={2020}
}

@article{shadrin2019analytical,
  title={Analytical review of standard Sae J3016 taxonomy and definitions for terms related to driving automation systems for on-road motor vehicles with latest updates},
  author={Shadrin, Sergej S and Ivanova, Anastasiia A},
  journal={Avtomobil'. Doroga. Infrastruktura.},
  number={3 (21)},
  pages={10},
  year={2019}
}

@inproceedings{capellier2018evidential,
  title={Evidential grid mapping, from asynchronous LIDAR scans and RGB images, for autonomous driving},
  author={Capellier, Edouard and Davoine, Franck and Fr{\'e}mont, Vincent and Iba{\~n}ez-Guzm{\'a}n, Javier and Li, You},
  booktitle={2018 21st International Conference on Intelligent Transportation Systems (ITSC)},
  pages={2595--2602},
  year={2018},
  organization={IEEE}
}

@inproceedings{lu2019l3,
  title={L3-net: Towards learning based lidar localization for autonomous driving},
  author={Lu, Weixin and Zhou, Yao and Wan, Guowei and Hou, Shenhua and Song, Shiyu},
  booktitle={Proceedings of the IEEE/CVF Conference on Computer Vision and Pattern Recognition},
  pages={6389--6398},
  year={2019}
}

@article{jaderberg2015spatial,
  title={Spatial transformer networks},
  author={Jaderberg, Max and Simonyan, Karen and Zisserman, Andrew and others},
  journal={Advances in neural information processing systems},
  volume={28},
  year={2015}
}

@inproceedings{he2016deep,
  title={Deep residual learning for image recognition},
  author={He, Kaiming and Zhang, Xiangyu and Ren, Shaoqing and Sun, Jian},
  booktitle={Proceedings of the IEEE conference on computer vision and pattern recognition},
  pages={770--778},
  year={2016}
}

@inproceedings{tan2019efficientnet,
  title={Efficientnet: Rethinking model scaling for convolutional neural networks},
  author={Tan, Mingxing and Le, Quoc},
  booktitle={International conference on machine learning},
  pages={6105--6114},
  year={2019},
  organization={PMLR}
}

@article{lateef2019survey,
  title={Survey on semantic segmentation using deep learning techniques},
  author={Lateef, Fahad and Ruichek, Yassine},
  journal={Neurocomputing},
  volume={338},
  pages={321--348},
  year={2019},
  publisher={Elsevier}
}

@article{hao2020brief,
  title={A brief survey on semantic segmentation with deep learning},
  author={Hao, Shijie and Zhou, Yuan and Guo, Yanrong},
  journal={Neurocomputing},
  volume={406},
  pages={302--321},
  year={2020},
  publisher={Elsevier}
}

@inproceedings{touvron2021training,
  title={Training data-efficient image transformers \& distillation through attention},
  author={Touvron, Hugo and Cord, Matthieu and Douze, Matthijs and Massa, Francisco and Sablayrolles, Alexandre and J{\'e}gou, Herv{\'e}},
  booktitle={International Conference on Machine Learning},
  pages={10347--10357},
  year={2021},
  organization={PMLR}
}

@article{xie2021segformer,
  title={SegFormer: Simple and efficient design for semantic segmentation with transformers},
  author={Xie, Enze and Wang, Wenhai and Yu, Zhiding and Anandkumar, Anima and Alvarez, Jose M and Luo, Ping},
  journal={Advances in Neural Information Processing Systems},
  volume={34},
  year={2021}
}

@article{zhu2020deformable,
  title={Deformable detr: Deformable transformers for end-to-end object detection},
  author={Zhu, Xizhou and Su, Weijie and Lu, Lewei and Li, Bin and Wang, Xiaogang and Dai, Jifeng},
  journal={arXiv preprint arXiv:2010.04159},
  year={2020}
}

@inproceedings{rashed2021generalized,
	title      = {{Generalized Object Detection on Fisheye Cameras for Autonomous Driving: Dataset, Representations and Baseline}},
	author     = {Rashed, Hazem and Mohamed, Eslam and Sistu, Ganesh and others},
	year       = 2021,
	booktitle  = {Proceedings of the IEEE/CVF Winter Conference on Applications of Computer Vision},
	pages      = {2272--2280}
}

@inproceedings{kumar2020unrectdepthnet,
	title      = {{UnRectDepthNet: Self-Supervised Monocular Depth Estimation using a Generic Framework for Handling Common Camera Distortion Models}},
	author     = {Ravi Kumar, Varun and Yogamani, Senthil and Bach, Markus and others},
	year       = 2020,
	booktitle  = {{IEEE/RSJ} International Conference on Intelligent Robots and Systems, {IROS} 2020, 2021},
	pages      = {8177--8183},
	doi        = {10.1109/iros45743.2020.9340732}
}

@inproceedings{uricar2021let,
	title      = {{Let's Get Dirty: GAN Based Data Augmentation for Camera Lens Soiling Detection in Autonomous Driving}},
	author     = {Uricar, Michal and Sistu, Ganesh and Rashed, Hazem and others},
	year       = 2021,
	booktitle  = {Proceedings of the IEEE/CVF Winter Conference on Applications of Computer Vision},
	pages      = {766--775}
}

@conference{chennupati2019auxnet,
	title      = {{AuxNet: Auxiliary Tasks Enhanced Semantic Segmentation for Automated Driving}},
	author     = {Sumanth Chennupati and Ganesh Sistu and Senthil Yogamani and others},
	year       = 2019,
	booktitle  = {Proceedings of the 14th International Joint Conference on Computer Vision, Imaging and Computer Graphics Theory and Applications (VISAPP)},
	pages      = {645--652}
}

@article{sistu2019real,
	title      = {{Real-time joint object detection and semantic segmentation network for automated driving}},
	author     = {Sistu, Ganesh and Leang, Isabelle and Yogamani, Senthil},
	journal    = {NeurIPS 2018 Workshop on ML on the Phone and other Consumer Devices},
	year       = {2018}
}

@inproceedings{uricar2019desoiling,
	title      = {{Desoiling dataset: Restoring soiled areas on automotive fisheye cameras}},
	author     = {Uric{\'a}r, Michal and Ulicny, Jan and Sistu, Ganesh and others},
	year       = 2019,
	booktitle  = {{IEEE/CVF} International Conference on Computer Vision Workshops, {ICCV} Workshops 2019},
	publisher  = {IEEE},
	pages      = {4273--4279}
}

@misc{rw2019timm,
  author = {Ross Wightman},
  title = {PyTorch Image Models},
  year = {2019},
  publisher = {GitHub},
  journal = {GitHub repository},
  doi = {10.5281/zenodo.4414861},
  howpublished = {\url{https://github.com/rwightman/pytorch-image-models}}
}

@article{roddick2018orthographic,
  title={Orthographic feature transform for monocular 3d object detection},
  author={Roddick, Thomas and Kendall, Alex and Cipolla, Roberto},
  journal={arXiv preprint arXiv:1811.08188},
  year={2018}
}

@inproceedings{Cordts2016Cityscapes,
title={The Cityscapes Dataset for Semantic Urban Scene Understanding},
author={Cordts, Marius and Omran, Mohamed and Ramos, Sebastian and Rehfeld, Timo and Enzweiler, Markus and Benenson, Rodrigo and Franke, Uwe and Roth, Stefan and Schiele, Bernt},
booktitle={Proc. of the IEEE Conference on Computer Vision and Pattern Recognition (CVPR)},
year={2016}
}
}
\end{document}